\documentclass[sigconf]{acmart}
\usepackage{tabularx}
\usepackage{bm}

\AtBeginDocument{%
  \providecommand\BibTeX{{%
    \normalfont B\kern-0.5em{\scshape i\kern-0.25em b}\kern-0.8em\TeX}}}

\setcopyright{rightsretained}

\acmSubmissionID{163}

\begin{document}

\title[When Does Pretraining Help?]{When Does Pretraining Help? Assessing Self-Supervised Learning for Law and the CaseHOLD Dataset of 53,000+ Legal Holdings}

\author{Lucia Zheng}
\authornote{These authors contributed equally to this work.}
\email{zlucia@stanford.edu}
\affiliation{%
 \institution{Stanford University}
 \city{Stanford}
 \state{California}
 \country{USA}
 }

\author{Neel Guha}
\authornotemark[1]
\email{nguha@stanford.edu}
\affiliation{%
 \institution{Stanford University}
 \city{Stanford}
 \state{California}
 \country{USA}
 }
 
\author{Brandon R. Anderson}
\email{banderson@law.stanford.edu}
\affiliation{%
 \institution{Stanford University}
 \city{Stanford}
 \state{California}
 \country{USA}
 }

\author{Peter Henderson}
\email{phend@stanford.edu}
\affiliation{%
 \institution{Stanford University}
 \city{Stanford}
 \state{California}
 \country{USA}
 }
 
\author{Daniel E. Ho}
\email{dho@law.stanford.edu}
\affiliation{%
 \institution{Stanford University}
 \city{Stanford}
 \state{California}
 \country{USA}
 }

\renewcommand{\shortauthors}{Zheng and Guha, et al.}

\begin{abstract}
While self-supervised learning has made rapid advances in natural language processing, it remains unclear when researchers should engage in resource-intensive domain-specific pretraining (domain pretraining). The law, puzzlingly, has yielded few documented instances of substantial gains to domain pretraining in spite of the fact that legal language is widely seen to be unique.  We hypothesize that these existing results stem from the fact that existing legal NLP tasks are too easy and fail to meet conditions for when domain pretraining can help.  To address this, we first present CaseHOLD (Case \underline{H}oldings \underline{O}n \underline{L}egal \underline{D}ecisions), a new dataset comprised of over 53,000+ multiple choice questions to identify the relevant holding of a cited case. This dataset presents a fundamental task to lawyers and is both legally meaningful and difficult from an NLP perspective (F1 of 0.4 with a BiLSTM baseline). Second, we assess performance gains on  CaseHOLD and existing legal NLP datasets.  While a Transformer architecture (BERT) pretrained on a general corpus (Google Books and Wikipedia) improves performance, domain pretraining (on a corpus of $\approx$3.5M decisions across all courts in the U.S. that is larger than BERT's) with a custom legal vocabulary exhibits the most substantial performance gains with CaseHOLD (gain of 7.2\% on F1, representing a 12\% improvement on BERT) and consistent performance gains across two other legal tasks. Third, we show that domain pretraining may be warranted when the task exhibits sufficient similarity to the pretraining corpus: the level of performance increase in three legal tasks was directly tied to the domain specificity of the task. Our findings inform when researchers should engage in resource-intensive pretraining and show that Transformer-based architectures, too, learn embeddings suggestive of distinct legal language.
\end{abstract}

\copyrightyear{2021}
\acmYear{2021}
\acmConference[ICAIL'21]{Eighteenth International Conference for Artificial Intelligence and Law}{June 21--25, 2021}{São Paulo, Brazil}
\acmBooktitle{Eighteenth International Conference for Artificial Intelligence and Law (ICAIL'21), June 21--25, 2021, São Paulo, Brazil}\acmDOI{10.1145/3462757.3466088}
\acmISBN{978-1-4503-8526-8/21/06}



\maketitle

\section{Introduction}
How can rapid advances in Transformer-based architectures be leveraged to address problems in law?  One of the most significant advances in natural language processing (NLP) has been the advent of ``pretrained'' (or self-supervised) language models, starting with Google's BERT model \cite{devlin-etal-2019-bert}. Such models are pretrained on a large corpus of general texts --- Google Books and Wikipedia articles --- resulting in significant gains on a wide range of fine-tuning tasks with much smaller datasets and have inspired a wide range of applications and extensions \cite{liu2019roberta, rogers2021primer}. 

One of the emerging puzzles for law has been that while \emph{general} pretraining (on the Google Books and Wikipedia corpus) boosts performance on a range of legal tasks, there do not appear to be any meaningful gains from \emph{domain-specific} pretraining (domain pretraining) using a corpus of law. Numerous studies have attempted to apply comparable Transformer architectures to pretrain language models on law, but have found marginal or insignificant gains on a range of legal tasks~\cite{chalkidis-etal-2020-legal, Elwany2019, zhong2020does, zhong2020jec}.  These results would seem to challenge a fundamental tenet of the legal profession: that legal language is \emph{distinct} in vocabulary, semantics, and reasoning \cite{mellinkoff2004language,mertz2007language, tiersma1999legal}.  Indeed, a common refrain for the first year of U.S.\ legal education is that students should learn the ``language of law'': ``Thinking like a lawyer turns out to
depend in important ways on speaking (and reading, and writing) like a lawyer.'' \cite{mertz2007language}.

We hypothesize that the puzzling failure to find substantial gains from domain pretraining in law stem from the fact that existing fine-tuning tasks may be too easy and/or fail to correspond to the domain of the pretraining corpus task. We show that  existing legal NLP tasks, Overruling (whether a sentence overrules a prior case, see Section \ref{sec:overruling}) and Terms of Service (classification of contractual terms of service, see Section \ref{sec:tos}), are simple enough for naive baselines (BiLSTM) or BERT (without domain-specific pretraining) to achieve high performance.  Observed gains from domain pretraining are hence relatively small. Because U.S.\ law lacks any benchmark task that is comparable to the large, rich, and challenging datasets that have fueled the general field of NLP (e.g., SQuAD~\cite{Rajpurkar2016}, GLUE~\cite{Wang2018}, CoQA~\cite{reddy2019coqa}), we present a new dataset that simulates a fundamental task for lawyers: identifying the legal \emph{holding} of a case. Holdings are central to the common law system. They represent the governing legal rule when the law is applied to a particular set of facts. The holding is precedential and what litigants can rely on in subsequent cases. So central is the identification of holdings that it forms a canonical task for first-year law students to identify, state, and reformulate the holding. 

This CaseHOLD dataset (Case \underline{H}oldings \underline{o}n \underline{L}egal \underline{D}ecisions) provides 53,000+ multiple choice questions with prompts from a judicial decision and multiple potential holdings, one of which is correct, that could be cited.  We construct this dataset using the rules of case citation \cite{bluebook}, which allow us to match a proposition to a source through a comprehensive corpus of U.S.\ case law from 1965 to the present. Intuitively, we extract all legal citations and use the ``holding statement,'' often provided in parenthetical propositions accompanying U.S.\ legal citations, to match context to holding \cite{arredondo2017harvesting}.

CaseHOLD extracts the context, legal citation, and holding statement and matches semantically similar, but inappropriate, holding propositions. This turns the identification of holding statements into a multiple choice task.

In Table \ref{tab:casehold}, we show a citation example from the CaseHOLD dataset. The Citing Text (prompt) consists of the context and legal citation text, Holding Statement 0 is the correct corresponding holding statement, Holding Statements 1-4 are the four similar, but incorrect holding statements matched with the given prompt, and the Label is the 0-index label of the correct holding statement answer. For simplicity, we use a fixed context window that may start mid-sentence.

\begin{table}[ht]
    \centering
    \small
    \caption{CaseHOLD example}
    \vspace{-0.15in}
    \begin{tabularx}{\columnwidth}{X}
         \toprule
         Citing Text (prompt) \\
         \midrule
         They also rely on Oswego Laborers' Local 214 Pension Fund v. Marine Midland Bank, 85 N.Y.2d 20, 623 N.Y.S.2d 529, 647 N.E.2d 741 (1996), which held that a plaintiff ``must demonstrate that the acts or practices have a broader impact on consumers at large." Defs.' Mem. at 14 (quoting Oswego Laborers', 623 N.Y.S.2d 529, 647 N.E.2d at 744). As explained above, however, Plaintiffs have adequately alleged that Defendants' unauthorized use of the DEL MONICO's name in connection with non-Ocinomled restaurants and products caused consumer harm or injury to the public, and that they had a broad impact on consumers at large inasmuch as such use was likely to cause consumer confusion. See, e.g., CommScope, Inc. of N.C. v. CommScope (U.S.A) Int'l Grp. Co., 809 F. Supp.2d 33, 38 (N.D.N.Y 2011) (\textless\text{HOLDING}\textgreater); New York City Triathlon, LLC v. NYC Triathlon \\
         \midrule
         Holding Statement 0 (correct answer) \\
         \midrule
         holding that plaintiff stated a 349 claim where plaintiff alleged facts plausibly suggesting that defendant intentionally registered its corporate name to be confusingly similar to plaintiffs CommScope trademark \\
         \midrule
         Holding Statement 1 (incorrect answer) \\
         \midrule
         holding that plaintiff stated a claim for breach of contract when it alleged the government failed to purchase insurance for plaintiff as agreed by contract \\
         \midrule
         Holding Statement 2 (incorrect answer) \\
         \midrule
         holding that the plaintiff stated a claim for tortious interference \\
         \midrule
         Holding Statement 3 (incorrect answer) \\
         \midrule
         holding that the plaintiff had not stated a claim for inducement to breach a contract where she had not alleged facts sufficient to show the existence of an enforceable underlying contract \\
         \midrule
         Holding Statement 4 (incorrect answer) \\
         \midrule
         holding plaintiff stated claim in his individual capacity \\
         \bottomrule
    \end{tabularx}
    \label{tab:casehold}
    \vspace{-0.15in}
\end{table}

We show that this task is difficult for conventional NLP approaches (BiLSTM F1 = 0.4 and BERT F1 = 0.6), even though law students and lawyers are able to solve the task at high accuracy. We then show that there are substantial and statistically significant performance gains from domain pretraining with a custom vocabulary (which we call Legal-BERT), using all available case law from 1965 to the present (a 7.2\% gain in F1, representing a 12\% relative boost from BERT). We then experimentally assess conditions for gains from domain pretraining with CaseHOLD and find that the size of the fine-tuning task is the principal other determinant of gains to domain-specific pretraining.

The code, the legal benchmark task datasets, and the Legal-BERT models presented here can be found at: \url{https://github.com/reglab/casehold}.

Our paper informs how researchers should decide when to engage in data and resource-intensive pretraining. Such decisions pose an important tradeoff, as cost estimates for fully pretraining BERT can be upward of \$1M  \cite{sharir2020cost}, with potential for social harm \cite{Ben:Geb:McM:21}, but advances in legal NLP may also alleviate huge disparities in access to justice in the U.S.\ legal system \cite{queudot2020improving, wu, engstrom}. Our findings suggest that there is indeed something unique to legal language when faced with sufficiently challenging forms of legal reasoning. 

\section{Related Work}
The Transformer-based language model, BERT \cite{devlin-etal-2019-bert}, which leverages a two step pretraining and fine-tuning framework, has achieved state-of-the-art performance on a diverse array of downstream NLP tasks.  BERT, however, was trained on a general corpus of Google Books and Wikipedia, and much of the scientific literature has since focused on the question of whether the Transformer-based approach could be improved by domain-specific pretraining. 

Outside of the law, for instance, Lee et al. \cite{lee_biobert_2019} show that BioBERT, a BERT model pretrained on biomedicine domain-specific corpora (PubMed abstracts and full text articles), can significantly outperform BERT on domain-specific biomedical NLP tasks.  For instance, it achieves gains of 6-9\% in strict accuracy compared to BERT \cite{lee_biobert_2019} for biomedical question answering tasks (BioASQ Task 5b and Task 5c) \cite{tsatsaronis_overview_2015}. Similarly, Beltagy et al. show improvements from domain pretraining with SciBERT, using a multi-domain corpus of scientific publications \cite{beltagy-etal-2019-scibert}.  On the ACL-ARC multiclass classification task \cite{jurgens-etal-2018-measuring}, which contains example citations labeled with one of six classes, where each class is a citation function (e.g., background), SciBERT achieves gains of 7.07\% in macro F1 \cite{beltagy-etal-2019-scibert}. It is worth noting that this task is constructed from citation text, making it comparable to the CaseHOLD task we introduce in Section \ref{sec:casehold}.

Yet work adapting this framework for the legal domain has not yielded comparable returns. Elwany at el. \cite{Elwany2019} use a proprietary corpus of legal agreements to pretrain BERT and report  ``marginal'' gains of 0.4 - 0.7\% on F1.  They note that in some settings, such gains could still be practically important.  Zhong et al. \cite{zhong2020does} uses BERT pretrained on Chinese legal documents and finds no gains relative to non-pretrained NLP baseline models (e.g., LSTM).  Similarly, \cite{zhong2020jec} finds that the same pretrained model performs poorly on a legal question and answer dataset.  

Hendrycks et al. \cite{hendrycks2021measuring} found that in zero-shot and few-shot settings, state-of-the-art models for question answering, GPT-3 and UnifiedQA, have lopsided performance across subjects, performing with near-random accuracy on subjects related to human values, such as law and morality, while performing up to 70\% accuracy on other subjects. This result motivated their attempt to create a better model for the multistate bar exam by further pretraining RoBERTa \cite{liu2019roberta}, a variant of BERT, on 1.6M cases from the Harvard Law Library case law corpus. They found that RoBERTa fine-tuned on the bar exam  task achieved 32.8\% test accuracy without domain pretraining and 36.1\% test accuracy with further domain pretraining. They conclude that while ``additional pretraining on relevant high quality text can help, it may not be enough to substantially
increase \dots performance.'' 
Hendrycks et al. \cite{hendrycks2021aligning} highlight that future research should especially aim to increase language model performance on tasks in subject areas such as law and moral reasoning since aligning future systems with human values and understanding of human approval/disapproval necessitates high performance on such subject specific tasks.

Chalkidis et al. \cite{chalkidis-etal-2020-legal} explored the effects of law pretraining using various strategies and evaluate on a broader range of legal NLP tasks. These strategies include (a) using BERT out of the box, which is trained on general domain corpora, (b) further pretraining BERT on legal corpora (referred to as LEGAL-BERT-FP), which is the method also used by Hendrycks et al. \cite{hendrycks2021measuring}, and (c) pretraining BERT from scratch on legal corpora (referred to as LEGAL-BERT-SC). Each of these models is then fine-tuned on the downstream task. They report that a LEGAL-BERT variant, in comparison to tuned BERT, achieves a 0.8\% improvement in F1 on a binary classification task derived from the ECHR-CASES dataset \cite{chalkidis-etal-2019-neural}, a 2.5\% improvement in F1 on the multi-label classification task derived from ECHR-CASES, and between a 1.1-1.8\% improvement in F1 on multi-label classification tasks derived from subsets of the CONTRACTS-NER dataset \cite{10.1145/3086512.3086515, chalkidis2019neural}.  These gains are small when considering the substantial data and computational requirements of domain pretraining. Indeed, Hendrycks et al. \cite{hendrycks2021measuring} concluded that the documented marginal difference does not warrant domain pretraining. 

This existing work raises important questions for law and artificial intelligence.  First, these results might be seen to challenge the widespread belief in the legal profession that legal language is distinct \cite{mellinkoff2004language,mertz2007language, tiersma1999legal}.  Second, one of the core challenges in the field is that unlike general NLP, which has thrived on large benchmark datasets (e.g., SQuAD~\cite{Rajpurkar2016}, GLUE~\cite{Wang2018}, CoQA~\cite{reddy2019coqa}), there are few large and publicly available legal benchmark tasks for U.S.\ law.  This is explained in part due to the expense of labeling decisions and challenges around compiling large sets of legal documents \cite{Pah134}, leading approaches above to rely on non-English datasets \cite{zhong2020does, zhong2020jec} or proprietary datasets \cite{Elwany2019}. Indeed, there may be a kind of selection bias in available legal NLP datasets, as they tend to reflect tasks that have been solved by methods often pre-dating the rise of self-supervised learning. Third, assessment standards vary substantially, providing little guidance to researchers on whether domain pretraining is worth the cost.  Studies vary, for instance, in whether BERT is retrained with custom vocabulary, which is particularly important in fields where terms of art can defy embeddings of general language models.  Moreover, some comparisons are between (a) BERT pretrained at 1M iterations and (b) domain-specific pretraining on top of BERT (e.g., 2M iterations) \cite{lee_biobert_2019}. Impressive gains might hence be confounded because the domain pretrained model simply has had more time to train. Fourth, legal language presents unique challenges in substantial part because of extensive and complicated system of legal citation. Work has shown that conventional tokenization that fails to account for the structure of legal citations can improperly present the legal text \cite{bommarito2018lexnlp}.  For instance, sentence boundary detection (critical for BERT's next sentence prediction pretraining task) may fail with legal citations containing complicated punctuation \cite{savelka2017sentence}. Just as using an in-domain tokenizer helps in multilingual settings \cite{rust2020good}, using a custom tokenizer should improve performance consistently for the ``language of law.'' Last, few have examined differences across the kinds of tasks where pretraining may be helpful.

We address these gaps for legal NLP by (a) contributing a new, large dataset with the task of identification of holding statements that comes directly from U.S.\ legal decisions, (b) assessing the conditions under which domain pretraining can help. 

\section{The CaseHOLD Dataset}
\label{sec:casehold}
We present the CaseHOLD dataset as a new benchmark dataset for U.S.\ law.  Holdings are, of course, central to the common law system. They represent the governing legal rule when the law is applied to a particular set of facts. The holding is what is precedential and what litigants can rely on in subsequent cases. So central is the identification of holdings that it forms a canonical task for first-year law students to identify, state, and reformulate the holding. Thus, as for a law student, the goal of this task is two-fold: (1) understand case names and their holdings; (2) understand how to re-frame the relevant holding of a case to back up the proceeding argument.

CaseHOLD is a multiple choice question answering task derived from legal citations in judicial rulings. The citing context from the judicial decision serves as the prompt for the question. The answer choices are holding statements derived from citations following text in a legal decision. There are five answer choices for each citing text. The correct answer is the holding statement that corresponds to the citing text. The four incorrect answers are other holding statements.

We construct this dataset from the Harvard Law Library case law corpus (In our analyses below, the dataset is constructed from the holdout dataset, so that no decision was used for pretraining Legal-BERT.).  We extract the holding statement from citations (parenthetical text that begins with ``holding'') as the correct answer and take the text before it as the citing text prompt. We insert a <HOLDING> token in the position of the citing text prompt where the holding statement was extracted. To select four incorrect answers for a citing text, we compute the TF-IDF similarity between the correct answer and the pool of other holding statements extracted from the corpus and select the most similar holding statements, to make the task more difficult. We set an upper threshold for similarity to rule out indistinguishable holding statements (here 0.75), which would make the task impossible. One of the virtues of this task setup is that we can easily tune the difficulty of the task by varying the context window, the number of potential answers, and the similarity thresholds. In future work, we aim to explore how modifying the thresholds and task difficulty affects results. In a human evaluation, the benchmark by a law student was an accuracy of 0.94.\footnote{This human benchmark was done on a pilot iteration of the benchmark dataset and may not correspond to the exact TF-IDF threshold presented here.} 

A full example of CaseHOLD consists of a citing text prompt, the correct holding statement answer, four incorrect holding statement answers, and a label 0-4 for the index of the correct answer. The ordering of indices of the correct and incorrect answers are random for each example and unlike a multi-class classification task, the answer indices can be thought of as multiple choice letters (A, B, C, D, E), which do not represent classes with underlying meaning, but instead just enumerate the answer choices. We provide a full example from the CaseHOLD dataset in Table \ref{tab:casehold}.

\section{Other Datasets}
\label{sec:datasets}

To provide a comparison on difficulty and domain specificity, we also rely on two other legal benchmark tasks. The three datasets are summarized in Table~\ref{tab:datasets}. 

\begin{table}[tb!]
    \centering
    \small
    \caption{Dataset overview}
    \vspace{-0.15in}
    \begin{tabularx}{\columnwidth}{XXcc}
         \toprule
         Dataset & Source & Task Type & Size \\
         \midrule
         Overruling & Casetext & Binary classification & 2,400 \\
         Terms of Service & Lippi et al.~\cite{Lippi_2019} & Binary classification & 9,414 \\
         CaseHOLD & Authors & Multiple choice QA & 53,137 \\
         \bottomrule
    \end{tabularx}
    \label{tab:datasets}
    \vspace{-0.15in}
\end{table}

In terms of size, publicly available legal tasks are small compared to mainstream NLP datasets (e.g., SQuAD has 100,000+ questions). The cost of obtaining high-fidelity labeled legal datasets is precisely why pretraining is appealing for law \cite{engstrom2020algorithmic}.  The Overruling dataset, for instance, required paying attorneys to label each individual sentence. Once a company has collected that information, it may not want to distribute it freely for the research community.  In the U.S.\ system, much of this meta-data is hence retained behind proprietary walls (e.g., Lexis and Westlaw), and the lack of large-scale U.S.\ legal NLP datasets has likely impeded scientific progress. 

We now provide more detail of the two other benchmark datasets. 

\subsection{Overruling}
\label{sec:overruling}
The Overruling task is a binary classification task, where positive examples are overruling sentences and negative examples are non-overruling sentences from the law. An overruling sentence is a statement that nullifies a previous case decision as a precedent, by a constitutionally valid statute or a decision by the same or higher ranking court which establishes a different rule on the point of law involved.

The Overruling task dataset was provided by Casetext, a company focused on legal research software. Casetext selected positive overruling samples through manual annotation by attorneys and negative samples through random sampling sentences from the Casetext law corpus. This procedure has a low false positive rate for negative samples because the prevalence of overruling sentences in the whole law is low. Less than 1\% of cases overrule another case and within those cases, usually only a single sentence contains overruling language. Casetext validates this procedure by estimating the rate of false positives on a subset of sentences randomly sampled from the corpus and extrapolating this rate for the whole set of randomly sampled sentences to determine the proportion of sampled sentences to be reviewed by human reviewers for quality assurance.

Overruling has moderate to high domain specificity because the positive and negative overruling examples are sampled from the Casetext law corpus, so the language in the examples is quite specific to the law. However, it is the easiest of the three legal benchmark tasks, since many overruling sentences are distinguishable from non-overruling sentences due to the specific and explicit language judges typically use when overruling. In his work on overruling language and speech act theory, Dunn cites several examples of judges employing an explicit performative form when overruling, using keywords such as ``overrule'', ``disapprove'', and ``explicitly reject'' in many cases \cite{dunn2003judges}. Language models, non-neural machine models, and even heuristics generally detect such keyword patterns effectively, so the structure of this task makes it less difficult compared to other tasks. Previous work has shown that SVM classifiers achieve high performance on similar tasks; Sulea et al. \cite{sulea:2017:legal} achieves a 96\% F1 on predicting case rulings of cases judged by the French Supreme Court and Aletras et al. \cite{aletras2016predicting} achieves 79\% accuracy on predicting judicial decisions of the European Court of Human Rights.

The Overruling task is important for lawyers because the process of verifying whether cases remain valid and have not been overruled is critical to ensuring the validity of legal arguments. This need has led to the broad adoption of proprietary systems, such as Shepard's (on Lexis Advance) and KeyCite (on Westlaw), which have become important legal research tools for most lawyers \cite{doi:10.1080/02703190802365671}. High language model performance on the Overruling tasks could enable further automation of the shepardizing process.

In Table \ref{tab:overruling}, we show a positive example of an overruling sentence and a negative example of a non-overruling sentence from the Overruling task dataset. Positive examples have label 1 and negative examples have label 0.

\begin{table}[htb]
    \centering
    \small
    \caption{Overruling examples}
    \vspace{-0.15in}
    \begin{tabularx}{\columnwidth}{Xc}
         \toprule
         Passage & Label \\
         \midrule
         for the reasons that follow, we approve the first district in the instant case and disapprove the decisions of the fourth district. & 1 \\
         \midrule
         a subsequent search of the vehicle revealed the presence of an additional syringe that had been hidden inside a purse located on the passenger side of the vehicle. & 0 \\
         \bottomrule
    \end{tabularx}
    \label{tab:overruling}
    \vspace{-0.2in}
\end{table}

\subsection{Terms of Service}
\label{sec:tos}
The Terms of Service task is a binary classification task, where positive examples are potentially unfair contractual terms (clauses) from the terms of service in contract documents. The Unfair Terms in Consumer Contracts Directive 93/13/EEC \cite{directive93/13} defines an unfair contractual term as follows. A contractual term is unfair if: (1) it has not been individually negotiated; and (2) contrary to the requirement of good faith, it causes a significant imbalance in the parties rights and obligations, to the detriment of the consumer.

The Terms of Service dataset comes from Lippi et al. \cite{Lippi_2019}, which studies machine learning and natural language approaches for automating the detection of potentially unfair clauses in online terms of service and implements a system called CLAUDETTE based on the results of the study. The dataset was constructed from a corpus of 50 online consumer contracts. Clauses were manually annotated as clearly fair, potentially unfair, and clearly unfair. Positive examples were taken to be potentially unfair or clearly unfair clauses and negative examples were taken to be clearly fair clauses to dichotomize the task. Lippi et al. \cite{Lippi_2019} also studies a multi-class setting in which each clause is additionally labeled according to one of eight categories of clause unfairness (e.g. limitation of liability). We focus on the more general setting where clauses are only labeled according to whether they encompass any type of unfairness.

Terms of Service has low domain specificity relative to the Overruling and CaseHOLD tasks because examples are drawn from the terms of service text in consumer contracts. Extensive contracting language may be less prevalent in the  Casetext and Harvard case law corpora, although contracts cases of course are. The Terms of Service task is moderately difficult. Excluding ensemble methods, the classifier that achieves highest F1 performance in the general setting of Lippi et al. \cite{Lippi_2019} is a single SVM exploiting bag-of-words features, which achieves a 76.9\% F1.

The Terms of Service task is useful for consumers, since automation of the detection of potentially unfair contractual terms could help consumers better understand the terms they agree to when signing a contract and make legal advice about unfair contracts more accessible and widely available for consumers seeking it. It could also help consumer protection organizations and agencies work more efficiently \cite{Lippi_2019}.

In Table \ref{tab:tos}, we show a positive example of a potentially unfair clause and a negative example of a fair clause from the Terms of Service dataset. Positive examples have label 1 and negative examples have label 0.

\begin{table}[htb]
    \centering
    \small
    \caption{Terms of Service examples}
    \vspace{-0.15in}
    \begin{tabularx}{\columnwidth}{Xc}
         \toprule
         Passage & Label \\
         \midrule
         occasionally we may, in our discretion, make changes to the agreements. & 1 \\
         \midrule
         this section contains service-specific terms that are in addition to the general terms. & 0 \\
         \bottomrule
    \end{tabularx}
    \label{tab:tos}
    \vspace{-0.15in}
\end{table}

\section{Methods}
\label{sec:methods}

Our basic approach to understand the conditions for when domain pretraining may help is to use a series of pretrained BERT models, but to carefully vary one key modeling decision at a time. This is computationally expensive requiring approximately 16 TPU (64 GPU) core-days per 1M steps. First, we assess performance with base BERT. Second, we train BERT with twice the number of iterations to be able to compare the value of additional training. Third, we ingest the entire Harvard Law case corpus from 1965 to the present and pretrain Legal-BERT on the corpus. The size of this dataset (37GB) is substantial, representing 3,446,187 legal decisions across all federal and state courts, and is larger than the size of the BookCorpus/Wikipedia corpus originally used to train BERT (15GB). Fourth, we train a custom vocabulary variant of Legal-BERT. We provide a comparison to a BiLSTM baseline. We now provide details of these methods.

\subsection{Baseline}
Our baseline architecture is a one-layer BiLSTM, with 300D word2vec vectors \cite{DBLP:journals/corr/abs-1301-3781}. For single-sentence tasks, Overruling and Terms of Service, we encode the sentence and pass the resulting vector to a softmax classifier. For CaseHOLD, each citation prompt has five answer choices associated with it. We concatenate the prompt with each one of the five answers, separated by the <SEP> token, to get five prompt-answer pairs. We independently encode each prompt-answer pair and pass the resulting vector through a linear layer, then apply softmax over the concatenated outputs for the five pairs. We choose this architecture because it is comparable to the design suggested for fine-tuning BERT on multiple choice tasks in Radford et al. \cite{radford2018improving}, where prompt-answer pairs are fed independently through BERT and a linear layer. In this architecture, we replace BERT with the BiLSTM.

\subsection{BERT}
We use the base BERT model (uncased, 110M parameters) \cite{devlin-etal-2019-bert} as our baseline BERT model. Because researchers in other disciplines have commonly performed domain pretraining starting with BERT's parameter values, we also train a model initialized with base BERT and pretrained for an additional 1M steps, using the same English Wikipedia corpus that BERT base was pretrained on. This facilitates a direct comparison to rule out gains solely from increased pretraining. We refer to this model, trained for 2M  total steps, as BERT (double), and compare it to our two Legal-BERT variants, each pretrained for 2M total steps. Using 2M steps as our comparison point for pretraining also allows us to address findings from Liu et al. \cite{liu2019roberta} that BERT was significantly undertrained and exhibited improved performance with RoBERTa, a set of modifications to BERT training procedure which includes pretraining the model longer.

\subsection{Legal-BERT}
We pretrain two variants of BERT with the Harvard Law case corpus (\url{https://case.law/}) from 1965 to the present.\footnote{We use this period because there is a significant change in the number of reporters around this period and it corresponds to the modern post-Civil Rights Act era.} We randomly sample 10\% of decisions from this corpus as a holdout set, which we use to create the CaseHOLD dataset. The remaining 90\% is used for pretraining. 

We preprocess the case law corpus with the sentence segmentation procedure and use the pretraining procedure described in Devlin et al. \cite{devlin-etal-2019-bert}. One variant is initialized with the BERT base model and pretrained for an additional 1M steps using the case law corpus and the same vocabulary as BERT (uncased). The other variant, which we refer to as Custom Legal-BERT, is pretrained from scratch for 2M steps using the case law corpus and has a custom legal domain-specific vocabulary. The vocabulary set is constructed using SentencePiece \cite{kudo2018sentencepiece} on a subsample (appx. 13M) of sentences from our pretraining corpus, with the number of tokens fixed to 32,000. We pretrain both variants with sequence length 128 for 90\% and sequence length 512 for 10\% over the 2M steps total.

Both Legal-BERT and Custom Legal-BERT are pretrained using the masked language model (MLM) pretraining objective, with whole word masking. Whole word masking and other knowledge masking strategies, like phrase-level and entity-level masking, have been shown to yield substantial improvements on various downstream NLP tasks for English and Chinese text, by making the MLM objective more challenging and enabling the model to learn more about prior knowledge through syntactic and semantic information extracted from these linguistically-informed language units \cite{cui2019pretraining, joshi-etal-2020-spanbert, sun2019ernie}. More recently, Kang et al. \cite{kang-etal-2020-neural} posit that whole-word masking may be most suitable for domain adaptation on emrQA \cite{pampari-etal-2018-emrqa}, a corpus for question answering on electronic medical records, because most words in emrQA are tokenized to sub-word WordPiece tokens \cite{wu2016googles} in base BERT due to the high frequency of unique, domain-specific medical terminologies that appear in emrQA, but are not in the base BERT vocabulary. Because the case law corpus shares this property of containing many domain-specific terms relevant to the law, which are likely tokenized into sub-words in base BERT, we chose to use whole word masking for pretraining the Legal-BERT variants on the legal domain-specific case law corpus. 

The second pretraining task is next sentence prediction. Here, we use regular expressions to ensure that legal citations are included as part of a segmented sentence according to the Bluebook system of legal citation \cite{bluebook}. Otherwise, the model could be poorly trained on improper sentence segmentation \cite{savelka2017sentence}.\footnote{Where the vagaries of legal citations create detectable errors in sentence segmentation (e.g., sentences with fewer than 3 words), we omit the sentence from the corpus.}

\section{Results}
\subsection{Base Setup}
After pretraining the models as described above in Section~\ref{sec:methods}, we fine-tune on the legal benchmark target tasks and evaluate the performance of each model.

\subsubsection{Hyperparameter Tuning}
We provide details on our hyperparameter tuning process at \url{https://github.com/reglab/casehold}.

\subsubsection{Fine-tuning and Evaluation}
For the BERT-based models, we use the input transformations described in Radford et al. \cite{radford2018improving} for fine-tuning BERT on classification and multiple choice tasks, which convert the inputs for the legal benchmark tasks into token sequences that can be processed by the pretrained model, followed by a linear layer and a softmax. For the CaseHOLD task, we avoid making extensive changes to the architecture used for the two classification tasks by converting inputs consisting of a prompt and five answers into five prompt-answer pairs (where the prompt and answer are separated by a delimiter token) that are each passed independently through our pretrained models followed by a linear layer, then take a softmax over the five concatenated outputs. For Overruling and Terms of Service, we use a single NVIDIA V100 (16GB) GPU to fine-tune on each task. For CaseHOLD, we used eight NVIDIA V100 (32GB) GPUs to fine-tune on the task.

We use 10-fold cross-validation to evaluate our models on each task. We use F1 score as our performance metric for the Overruling and Terms of Service tasks and macro F1 score as our performance metric for CaseHOLD, reporting mean F1 scores over 10 folds. We report our model performance results in Table \ref{tab:base_performance} and report statistical significance from (paired) $t$-tests with 10 folds of the test data to account for uncertainty.

\begin{table*}[ht!]
    \centering
    \caption{Test performance, with $\pm 1.96 \times \text{standard error}$, aggregated across 10 folds. Mean F1 scores are reported for Overruling and Terms of Service. Mean macro F1 scores are reported for CaseHOLD. The best scores are in bold.}
    \begin{tabular}{cccccc}
        \toprule
        Model & Baseline & BERT & BERT (double) & Legal-BERT & Custom Legal-BERT \\
        \midrule
        Overruling --- \textcolor{ACMDarkBlue}{DS=-0.028} & $0.910 \pm 0.012$ & $0.958 \pm 0.005$ & $0.958 \pm 0.005$ & $0.963 \pm 0.007$ & $\bm{0.974} \pm 0.005$ \\
        Terms of Service --- \textcolor{ACMDarkBlue}{DS=-0.085} & $0.712 \pm 0.020$ & $0.722 \pm 0.015$ & $0.773 \pm 0.019$ & $0.750 \pm 0.018$ & $\bm{0.787} \pm 0.013$ \\
        CaseHOLD --- \textcolor{ACMDarkBlue}{DS=0.084} & $0.399 \pm 0.005$ & $0.613 \pm 0.005$ & $0.623 \pm 0.003$ & $0.680 \pm 0.003$ & $\bm{0.695} \pm 0.003$ \\
        \midrule
        Number of Pretraining Steps & - & 1M & 2M & 2M & 2M \\
        \midrule
        Vocabulary Size (domain) & - & 30,522 (general) & 30,522 (general) & 30,522 (general) & 32,000 (legal) \\
        \bottomrule
    \end{tabular}
    \label{tab:base_performance}
\end{table*}

From the results of the base setup, for the easiest Overruling task, the difference in F1 between BERT (double) and Legal-BERT is 0.5\% and BERT (double) and Custom Legal-BERT is 1.6\%. Both of these differences are marginal. For the task with intermediate difficulty, Terms of Service, we find that BERT (double) with further pretraining BERT on the general domain corpus increases performance over base BERT by 5.1\%, but the Legal-BERT variants with domain-specific pretraining do not outperform BERT (double) substantially. This is likely because Terms of Service has low domain-specificity, so pretraining on legal domain-specific text does not help the model learn information that is highly relevant to the task. We note that BERT (double), with 77.3\% F1, and Custom Legal-BERT, with 78.7\% F1, outperform the highest performing model from Lippi et al. \cite{Lippi_2019} for the general setting of Terms of Service, by 0.4\% and 1.8\% respectively. For the most difficult and domain-specific task, CaseHOLD, we find that Legal-BERT and Custom Legal-BERT both substantially outperform BERT (double) with gains of 5.7\% and 7.2\% respectively. Custom Legal-BERT achieves the highest F1 performance for CaseHOLD, with a macro F1 of 69.5\%.

We run paired $t$-tests to validate the statistical significance of model performance differences for a 95\% confidence interval. The mean differences between F1 for paired folds of BERT (double) and base BERT are statistically significant for the Terms of Service task, with $p$-value $<0.001$. Additionally, the mean differences between F1 for paired folds of Legal-BERT and BERT (double) with $p$-value $<0.001$ and the mean differences between F1 for paired folds of Custom Legal-BERT and BERT (double) with $p$-value $<0.001$ are statistically significant for the CaseHOLD task. The substantial performance gains from the Legal-BERT model variants were achieved likely because the CaseHOLD task is adequately difficult and highly domain-specific in terms of language.

\subsubsection{Domain Specificity Score}
Table \ref{tab:base_performance} also provides a measure of domain specificity of each task, which we refer to as the domain specificity (DS) score. We define DS score as the average difference in pretrain loss between Legal-BERT and BERT, evaluated on the downstream task of interest. For a specific example, we run prediction for the downstream task of interest on the example input using Legal-BERT and BERT models after pretraining, but before fine-tuning, calculate loss on the task (i.e., binary cross entropy loss for Overruling and Terms of Service, categorical cross entropy loss for CaseHOLD), and take the difference between the loss of the two models. Intuitively, when the difference is large, the general corpus does not predict legal language very well. DS scores serve as a heuristic for task domain specificity. A positive value conveys that on average, Legal-BERT is able to reason more accurately about the task compared to base BERT after the pretraining phase, but before fine-tuning, which implies the task has higher legal domain-specificity.

The rank order from least to most domain-specific is: Terms of Service, Overruling, and CaseHOLD. This relative ordering makes substantive sense. CaseHOLD has high domain specificity since a holding articulates a court's precise, legal statement of the holding of a decision.  As noted earlier, the language of contractual terms-of-service may not be represented extensively in the case law corpus.

The results in Table \ref{tab:base_performance} outline an increasing relationship between the legal domain specificity of a task, as measured by the DS score (compatible with our qualitative assessments of the tasks), and the degree to which prior legal knowledge captured by the model through unsupervised pretraining improves performance. Additionally, the Overruling results suggest that there exists an interplay between the legal domain specificity of a task and the difficulty of the task, as measured by baseline performance on non-attention based models. Gains from attention based models and domain pretraining may be limited for lower difficulty tasks, even those with intermediate DS scores, such as Overruling, likely because the task is easy enough provided local context that increased model domain awareness is only marginally beneficial.

\subsection{Task Variants}
To provide a further assessment on the conditions for pretraining, we evaluate the performance and sensitivity of our models on three task variants of CaseHOLD, the task for which we observe the most substantial gains from domain pretraining. We vary the task on three dimensions: the volume of training data available for fine-tuning (train volume), the difficulty of the prompt as controlled by the length of the prompt (prompt difficulty), and the level of domain specificity of the prompt (domain match). We hypothesize that these dimensions --- data volume, prompt difficulty, and domain specificity --- capture the considerations practitioners must account for in considering whether pretraining is beneficial for their use case. For the task variants, we split the CaseHOLD task dataset into three train and test set folds using an 80/20 split over three random seeds and evaluate on each fold. We report results as the mean F1 over the three folds' test sets.

\subsubsection{Train Volume}
For the train volume variant, keeping the test set constant, we vary the train set size to be of size 1, 10, 100, 500, 1,000, 5,000, 10,000, and the full train set. We find that the Legal-BERT gains compared to BERT (double) are strongest with low train volume and wash out with high train volume. As we expect, Legal-BERT gains are larger when the fine-tuning dataset is smaller. In settings with limited training data, the models must rely more on prior knowledge and Legal-BERT's prior knowledge is more relevant to the highly domain-specific task due to pretraining on legal domain-specific text, so we see stronger gains from Legal-BERT compared to BERT (double). For a training set size of 1, the mean gain in Legal-BERT is $17.6\% \pm 3.73$, the maximal gain across train set sizes.

This particular variant is well-motivated because it has often been challenging to adapt NLP for law precisely because there is limited labeled training data available. Legal texts typically require specialized legal knowledge to annotate, so it can often be prohibitively expensive to construct large structured datasets for the legal domain \cite{engstrom}.

\begin{figure}
    \centering
    \includegraphics[width=0.7\columnwidth]{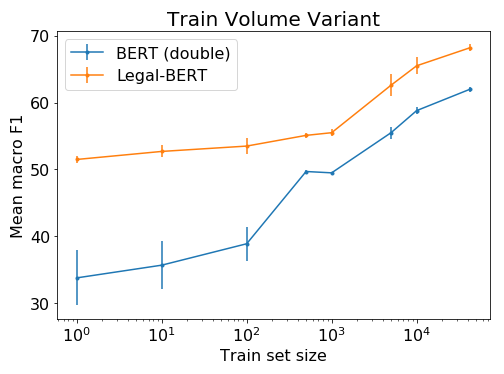}
    \vspace{-0.15in}
    \caption{Mean macro F1 scores over 3 folds, with $\pm 1.96 \times \text{standard error}$, for train volume variant.}
    \label{fig:trainvolume}
    \vspace{-0.15in}
\end{figure}

\subsubsection{Prompt Difficulty}
For the difficulty variant, we vary the citing text prompt difficulty, by shortening the length of the prompt to the first $x$ words. The average length of a prompt in the CaseHOLD task dataset is 136 words, so we take the first $x =$ 5, 10, 20, 40, 60, 80, 100 words of the prompt and the full prompt.  We take the first $x$ words instead of the last $x$ words closest to the holding, as the latter could remove less relevant context further from the holding and thus make the task easier. We find that the prompt difficulty variant does not result in a clear pattern of increasing gains from Legal-BERT over BERT (double) above 20 words, though we would expect to see the gains grow as the prompt is altered more. However, a 2\% drop in gain is seen in the 5 word prompt (the average F1 gap above 20 words is 0.062, while at 5 is 0.0391).

\begin{figure}
    \centering
    \includegraphics[width=0.7\columnwidth]{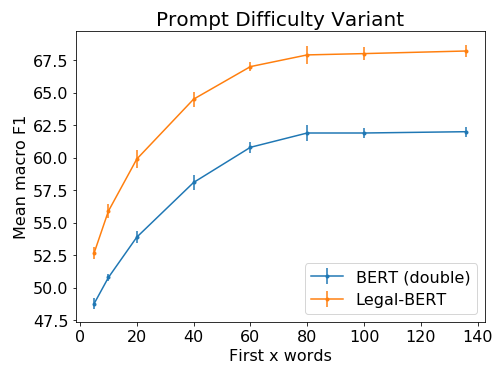}
    \vspace{-0.15in}
    \caption{Mean macro F1 scores over 3 folds, with $\pm1.96 \times \textbf{standard error}$, for prompt difficulty variant.}
    \label{fig:promptdifficulty}
    \vspace{-0.15in}
\end{figure}

One possible reason we do not observe a clear pattern may be that the baseline prompt length  constrains the degree to which we can manipulate the prompt and vary this dimension; the expected relationship may be more clearly observed for a dataset with longer prompts. Additionally, BERT models are known to disregard word order~\cite{sinha2020unnatural}. It is possible that beyond 5 words, there is a high likelihood that a key word or phrase is encountered that Legal-BERT has seen in the pretraining data and can attend to.

\subsubsection{Domain Match}
For the domain match variant, we weight the predictions for the test set when calculating F1 by sorting the examples in ascending order by their DS score and weighting each example by its rank order. Intuitively, this means the weighted F1 score rewards correct predictions on examples with higher domain specificity more. This method allows us to keep train volume constant, to avoid changing the domain specificity distribution of train set examples (which would occur if the test set was restricted to a certain range of DS scores), and still observe the effects of domain specificity on performance in the test set. We expect that the gains in Legal-BERT compared to BERT are stronger for the weighted F1 than the unweighted F1. We find that the mean gain in Legal-BERT over three folds is greater for the weighted F1 compared to the unweighted F1, but only by a difference of $0.8\% \pm 0.154$, as shown in Table \ref{tab:domainmatch}.

\begin{table}[ht]
    \centering
    \small
    \caption{Mean gain in Legal-BERT over 3 folds, for domain match variant.}
    \vspace{-0.15in}
    \begin{tabular}{cccc}
         \toprule
         Mean macro F1 & BERT (double) & Legal-BERT & Mean Gain \\
         \midrule
         Unweighted & 0.620 & 0.679 & 0.059 \\
         Weighted & 0.717 & 0.784 & 0.067 \\
         \bottomrule
    \end{tabular}
    \label{tab:domainmatch}
\end{table}

\begin{figure}
    \centering
    \includegraphics[width=\columnwidth]{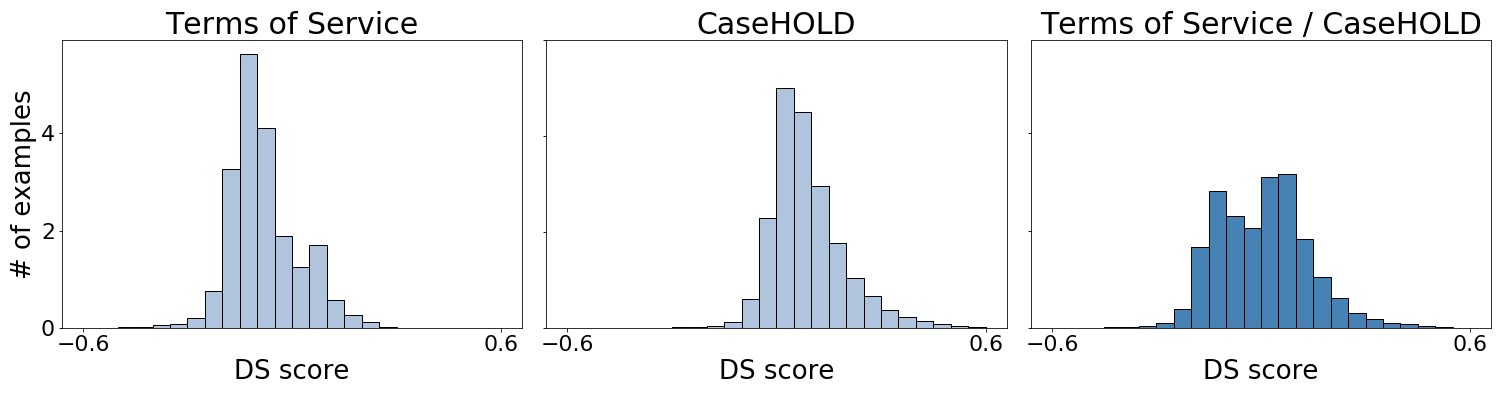}
    \vspace{-0.15in}
    \caption{Density histograms of DS scores of examples for Terms of Service, CaseHOLD, and both tasks.}
    \label{fig:domainmatch}
    \vspace{-0.15in}
\end{figure}

One possible reason this occurs is that the range of DS scores across examples in the CaseHOLD task is relatively small, so some similarly domain-specific examples may have fairly different rank-based weights. In Figure \ref{fig:domainmatch}, we show histograms of the DS scores of examples for Terms of Service, CaseHOLD, and 5,000 examples sampled (without replacement) from each task. Notice that the Terms of Service examples are skewed towards negative DS scores and the CaseHOLD examples are skewed towards positive DS scores so the range of DS scores within a task is limited, while the examples sampled from both tasks span a larger range, explaining the small gains from the domain match variant, but more substantial gains for CaseHOLD from Legal-BERT compared to Terms of Service. In other words, because the CaseHOLD task is already quite domain specific, variation \emph{within} the corpus may be too range-restricted to provide a meaningful test of domain match.

Further work could instead examine domain match by pretraining on specific areas of law (e.g., civil law) and fine-tuning on other areas (e.g., criminal law), but the Harvard case law corpus does not currently have meaningful case / issue type features.

\subsection{Error Analysis}
We engage in a brief error analysis by comparing the the breakdown of errors between Legal-BERT and BERT (double). In the test set, the breakdown was: 55\% both correct, 13\% Legal-BERT correct / BERT (double) incorrect, 7\% BERT (double) correct / Legal-BERT incorrect, and 25\% both incorrect. We read  samples of instances where model predictions diverged, with a focus on the examples Legal-BERT predicted correctly and BERT (double) predicted incorrectly. While we noticed instances indicative of Legal-BERT attending to legal language (e.g., identifying a different holding because of the difference between a ``may" and ``must" and because the ``but see" citation signal indicated a negation), we did not find that such simple phrases predicted differences in performance in a bivariate probit analysis.  We believe there is much fruitful work to  be done on further understanding what Legal-BERT uniquely attends to.

\subsection{Limitations}
While none of the CaseHOLD cases exist in the pretraining dataset, some of Legal-BERT gains on the CaseHOLD task may be attributable to having seen key words tied to similar holding formulations in the pretraining data. As mentioned, this is part of the goal of the task: understanding the holdings of important cases in a minimally labeled way and determining how the preceding context may affect the holding. This would explain the varying results in the prompt difficulty variant of the CaseHOLD task: gains could be mainly coming from attending to only a key word (e.g., case name) in the context. This may also explain how Legal-BERT is able to achieve zero-shot gains in the train volume variant of the task. 
BERT, may have also seen some of the cases and holdings in English Wikipedia,\footnote{See, e.g., \url{https://en.wikipedia.org/wiki/List_of_landmark_court_decisions_in_the_United_States} which contains a list of cases and their holdings.} potentially explaining its zero-shot performance improvements over random in the train volume variant.
Future work on the CaseHOLD dataset may wish to disentangle memorization of case names from the framing of the citing text, but we provide a strong baseline here. One possible mechanism for this is via a future variant of the CaseHOLD task where a case holding is paraphrased to indicate bias toward a different viewpoint from the contextual framing. This would reflect the first-year law student exercise of re-framing a holding to persuasively match their argument and isolate the two goals of the task.

\balance
\section{Discussion}
Our results resolve an emerging puzzle in legal NLP: if legal language is so unique, why have we seen only marginal gains to domain pretraining in law?  Our evidence suggests that these results can be explained by the fact that existing legal NLP benchmark tasks are either too easy or not domain matched to the pretraining corpus.  Our paper shows the largest gains documented for any legal task from pretraining, comparable to the largest gains reported by SciBERT and BioBERT \cite{beltagy-etal-2019-scibert, lee_biobert_2019}. Our paper also shows the highest performance documented for the general setting of the Terms of Service task \cite{Lippi_2019}, suggesting substantial gains from domain pretraining and tokenization.

Using a range of legal language tasks that vary in difficulty and domain-specificity, we find BERT already achieves high performance for easy tasks, so that further domain pretraining adds little value. For the intermediate difficulty task that is not highly domain-specific, domain pretraining can help, but gain is most substantial for highly difficult and domain-specific tasks.

These results suggest important future research directions.  First, we hope that the new CaseHOLD dataset will spark interest in solving the challenging environment of legal decisions.  Not only are many available benchmark datasets small or unavailable, but they may also be biased toward solvable tasks.  After all, a company would not invest in the Overruling task (baseline F1 with BiLSTM of 0.91), without assurance that there are significant gains to paying attorneys to label the data.  Our results show that domain pretraining may enable a much wider range of legal tasks to be solved.

Second, while the creation of large legal NLP datasets is impeded by the sheer cost of attorney labeling, CaseHOLD also illustrates an advantage of leveraging domain knowledge for the construction of legal NLP datasets.  Conventional segmentation would fail to take advantage of the complex system of legal citation, but investing in such preprocessing enables better representation and extraction of legal texts.

Third, our research provides guidance for researchers on when pretraining may be appropriate.  Such guidance is sorely needed, given the significant costs of language models, with one estimate suggesting that full  pretraining of BERT with a 15GB corpus can exceed \$1M.  Deciding whether to pretrain itself can hence have  significant ethical, social, and environmental implications \cite{Ben:Geb:McM:21}. Our research suggests that many easy tasks in law may not require domain pretraining, but that gains are most likely when ground truth labels are scarce and the task is sufficiently in-domain. Because estimates of domain-specificity across tasks using DS score match our qualitative understanding, this heuristic can also be deployed to determine whether pretraining is worth it. Our results suggest that for other high DS and adequately difficult legal tasks, experimentation with custom, task relevant approaches, such as leveraging corpora from task-specific domains and applying tokenization / sentence segmentation tailored to the characteristics of in-domain text, may yield substantial gains. Bender et al. \cite{Ben:Geb:McM:21} discuss the significant environmental costs associated in particular with transferring an existing large language models to a new task or developing new models, since these workflows require retraining to experiment with different model architectures and hyperparameters. DS scores provide a quick metric for future practitioners to evaluate when resource intensive model adaptation and experimentation may be warranted on other legal tasks. DS scores may also be readily extended to estimate the domain-specificity of tasks in other domains with existing pretrained models like SciBERT and BioBERT \cite{beltagy-etal-2019-scibert, lee_biobert_2019}.

In sum, we have shown that a new benchmark task, the CaseHOLD dataset, and a comprehensively pretrained Legal-BERT model illustrate the conditions for domain pretraining and suggests that language models, too, can embed what may be unique to legal language. 

\begin{acks}
    We thank Devshi Mehrotra and Amit Seru for research assistance, Casetext for the Overruling dataset, Stanford's Institute for Human-Centered Artificial Intelligence (HAI) and Amazon Web Services (AWS) for cloud computing research credits, and Pablo Arredondo, Matthias Grabmair, Urvashi Khandelwal, Christopher Manning, and Javed Qadrud-Din for helpful comments. 
\end{acks}

\bibliographystyle{ACM-Reference-Format}
\bibliography{legal-benchmarks}


\end{document}